\documentclass[conference]{IEEEtran}
\IEEEoverridecommandlockouts
\usepackage{cite}
\usepackage{amsmath,amssymb,amsfonts}
\usepackage{algorithmic}
\usepackage{graphicx}
\usepackage{textcomp}
\usepackage{xcolor}
\usepackage{url}
\usepackage{xurl}
\def\BibTeX{{\rm B\kern-.05em{\sc i\kern-.025em b}\kern-.08em
    T\kern-.1667em\lower.7ex\hbox{E}\kern-.125emX}}
\begin{document}

\title{AI on the Road: A Comprehensive Analysis of Traffic Accidents and  Accident Detection System in Smart Cities\\
}
\author{\IEEEauthorblockN{Anonymous for Review}
\IEEEauthorblockA{\textit{Anonymous for Review} \\
\textit{Anonymous for Review}\\
Anonymous for Review\\
Anonymous for Review}
\author{\IEEEauthorblockN{Victor Adewopo}
\IEEEauthorblockA{\textit{School of Information Technology} \\
\textit{University of Cincinnati}\\
Cincinnati, USA \\
Adewopva@mail.uc.edu}
\and
\IEEEauthorblockN{Nelly Elsayed}
\IEEEauthorblockA{\textit{School of Information Technology} \\
\textit{University of Cincinnati}\\
Cincinnati, USA \\
elsayeny@ucmail.uc.edu}
\and
\IEEEauthorblockN{Zag Elsayed}
\IEEEauthorblockA{\textit{School of Information Technology} \\
\textit{University of Cincinnati}\\
Cincinnati, USA \\
elsayezs@ucmail.uc.edu}
\and
\IEEEauthorblockN{Murat Ozer}
\IEEEauthorblockA{\textit{School of Information Technology} \\
\textit{University of Cincinnati}\\
Cincinnati, USA \\
ozermm@ucmail.uc.edu}
\and
\IEEEauthorblockN{Victoria Wangia-Anderson}
\IEEEauthorblockA{\textit{College of Allied Health Sciences} \\
\textit{University of Cincinnati}\\
Cincinnati, USA \\
wangiava@ucmail.uc.edu}
\and
\IEEEauthorblockN{Ahmed Abdelgawad}
\IEEEauthorblockA{\textit{College of Science and Engineering} \\
\textit{Central Michigan University}\\
Michigan, USA \\
abdel1a@cmich.edu}
}
}

\maketitle

\begin{abstract}
Accident detection and traffic analysis is a critical component of smart city and autonomous transportation systems that can reduce accident frequency, severity and improve overall traffic management. This paper presents a comprehensive analysis of traffic accidents in different regions across the United States using data from the National Highway Traffic Safety Administration (NHTSA) Crash Report Sampling System (CRSS). To address the challenges of accident detection and traffic analysis, this paper proposes a framework that uses traffic surveillance cameras and action recognition systems to detect and respond to traffic accidents spontaneously. Integrating the proposed framework with emergency services will harness the power of traffic cameras and machine learning algorithms to create an efficient solution for responding to traffic accidents and reducing human errors. Advanced intelligence technologies, such as the proposed accident detection systems in smart cities, will improve traffic management and traffic accident severity. Overall, this study provides valuable insights into traffic accidents in the US and presents a practical solution to enhance the safety and efficiency of transportation systems.
\end{abstract}


\section{Introduction}
Road traffic accidents are a significant public health concern, accounting for many fatalities and injuries globally each year \cite{nhtsa}. World Health Organization (WHO) report approximated that 1.35 million people die annually from road traffic accidents, and over 50 million suffer non-fatal injuries in road accidents~\cite {chang2020global}. Road accidents have significant financial implications and economic losses due to loss in productivity, medical expenses, and property and infrastructure damage. Adequate steps are yet to be deployed globally to address the catastrophic effects of road traffic accidents on environmental safety and population health. Some measure put in place in developing countries includes wearing helmets, enforcing speed limits, and penalizing traffic offenders. Despite efforts to reduce the incidence of road traffic fatalities, low-income countries are impacted mainly by the rise of road accidents due to the lack of infrastructure and emergency services in such events. While high-income countries have seen slow progress in reducing road traffic accidents over the past decade,  reports still indicate that 80\% of traffic accidents occur in developing countries \cite{mohammed2019review}. The adoption of the United Nations Decade of Action for Road Safety (2021-2030) in improving global road safety to prevent road traffic deaths and injuries by 50\% presents an opportunity to address this problem \cite{mohammed2019review}. Other research works have demonstrated the ability to reduce road accident fatality significantly by 40\% and save over 200,000 lives in post-crash care through evidence-based risk factor intervention \cite{vecino2022saving,sauerzapf2012road}. Andres et al.~\cite{vecino2022saving} compiled country-specific variables in over 185 countries in providing road traffic safety risk factors. Their research demonstrates that effective road safety measures can reduce the incidence of road traffic fatalities across all countries. 

In recent years, technological advancements have led to the development of autonomous accident detection systems. These systems are designed to detect accidents rapidly, which can reduce emergency response times, remove reporting errors, and save more lives \cite{Adewopo2022ReviewOA}. Based on the increasing popularity of these systems, there is growing interest in utilizing cost-effective systems in monitoring and detecting traffic accidents \cite{banstola2017cost}.

The effectiveness of these systems remains to be determined, despite their potential benefits~\cite{bhalla2019building}. In developing and implementing these systems, there are several challenges, including a lack of standardized approaches to accident detection, varying quality of the systems, and the need for data sharing and collaboration.
This paper aims to provide an overview of the impact of autonomous accident detection systems, analyze accident hotspots and analyze the secondary effects of traffic accidents on human and economic costs related to traffic accidents. Furthermore, this paper proposed a novel framework for detecting accidents based on an action recognition system.

\section{Current Accident Detection Techniques}
Traffic accidents are a significant cause of death and injury worldwide \cite{mohammed2019review}. In recent years, various techniques have been developed to improve traffic safety and reduce the number of accidents. One such technique is using sensors for traffic monitoring and accident detection. These sensors can detect changes in traffic flow, detect accidents, and even provide data for predicting future traffic conditions \cite{Khalil, Adewopo2022ReviewOA}. 
Khalil et al.~\cite{Khalil} proposed using ultrasonic sensors for automatic road accident detection. The proposed system consists of two ultrasonic sensors for measuring distance and sound waves for detecting collisions with obstacles. The system measures the distance between the sensor and the obstacle to detect accidents. Using autonomous vehicles and intelligent transportation systems in smart cities can reduce accidents and enhance road safety. Most techniques adopted for prompt monitoring of road accidents are expensive and sophisticated. The National Safety Council reported a 19\% increase in car fatality in some states across the US in the first half of 2022 as compared to 2021 \cite{NSC}. The World Health Organization (WHO) reports that 1.35 million people die yearly in road accidents, and about 50 million people get injured \cite{Khalil}. Modern-day technology such as surveillance cameras, Global Positioning Systems (GPS), Edge AI, and IoT can be beneficial in developing and deploying deep learning algorithms for detecting and predicting accidents. 
Accident detection is a crucial area of research in transportation engineering and road safety. With the rapid advancement in technology, various methods have been proposed for accident detection, ranging from traditional human reporting to modern automated systems that utilize sensors and computer vision. This section will discuss some of the current literature and methodologies used for accident detection.

\subsection{Human Reporting}
In order to process information, humans rely on three fallible mental functions; perception, attention, and memory. There is often a misconception that human error is responsible for accidents. However, the situation is often beyond the driver's ability to react \cite{green2004human}. Human reporting has been an essential source of accident data for decades. People involved in accidents or witnessing accidents can provide valuable information that can be used to develop accident prevention strategies. In this method, accidents are reported by eyewitnesses or other individuals who witness the accident. However, depending solely on human reporting can be unreliable and time-consuming. As a result, it may not be suitable for time-critical accident detection. Therefore, using computer vision techniques to detect accidents in real-time automatically is an area of active research. These computer vision systems use cameras and machine learning algorithms to analyze video data and detect accidents in real-time. The combination of sensors, IoT devices, human reporting, and computer vision has the potential to significantly improve traffic safety and accident prevention \cite{Mateen, Adewopo2022ReviewOA}. 

\subsection{Sensors}
Sensors such as accelerometers, GPS, and gyroscopes can be installed in vehicles to monitor driving behavior and detect accidents. These sensors can detect sudden changes in speed, direction, or orientation, which may indicate an accident. The data collected from these sensors can be processed by algorithms to detect and classify accidents \cite{Khalil}. Several studies have explored different types of sensors for accident detection. Syedul et al. \cite{Syedul} proposed using a GPS receiver to monitor the speed of a vehicle and detect accidents based on the monitored speed. 
Communications (GSM) \cite{Syedul}. 
White et al. \cite{white2011wreckwatch} research combines sensors and context data for accident detection, relying on smartphone sensors and web services to provide situational reports in traffic accidents to reduce the time taken to alert first responders about traffic accidents. Some potential limitation of this system includes the high power consumption of smartphone traffic accidents system, which may result in reduced battery life. The increased possibility of false alarms due to smartphone filters was also identified as a limitation for minor accidents such as sideswipe collisions when a car is in low-speed traffic \cite{white2011wreckwatch}. Other studies have proposed using more specific sensors, such as airbag systems, for accident detection. However, there is a need for a more generic system that can be utilized across all types of vehicles \cite{Khalil}. To address the limitations of previous studies, Mateen et al. \cite{Mateen} proposed a framework for an Accident Alert and Light and Sound  System (AALS) using pre-saved locations to reduce the time needed to locate accidents without relying on GPS. The framework attempts to reduce the risk of Multiple vehicle collisions by providing real-time notifications to nearby vehicles and emergency services. Using various sensors for accident detection is a promising area of research that can improve the safety of roads and reduce the occurrence of motor vehicle collisions (MVCs).

\subsection{Social Network and Geosocial Media Data} 
The enormous amount of information being constantly shared daily across various social media platforms contains artifacts that can be analyzed to generate meaningful insights for traffic events~\cite{rashidi2017exploring}. Although the ability to monitor and analyze the exploding information manually seems impossible based on the high volume and unstructured formats of the information being presented~\cite{adewopo2022deep}. Monitoring traffic-related information on social media has been proven beneficial in detecting traffic events.
Xu et al.~\cite{xu2018sensing} provided a synthesis of research work that explored the usage of geosocial media data for detecting traffic events. Events such as road accidents, road closures, and traffic conditions are typically shared among people through social media platforms. Such events can be tracked down with the aid of GPS in getting first responders to the event location and often contain information that triggered such events. 
Xu et al.~\cite{xu2019traffic} utilized Twitter data for mining and filtering noisy data by association rules among words related to traffic events. The proposed framework achieved 81\% accuracy in classifying data into non-traffic events, traffic accidents, roadwork, and severe weather conditions. Similarly, Salas et al.~\cite{Salas2017TrafficED} developed a framework leveraging social media data to crawl, process, and filter social media data for implying traffic incidents and real-time detection of traffic events with text classification algorithm~\cite{gu2016twitter}.

\section{Proliferation of Traffic Accident Event in the United States}
To understand factors contributing to traffic accidents and develop potential strategies to improve highway safety, we analyze traffic accidents using data from the National Highway Traffic Safety Administration (NHTSA) Crash Report Sampling System (CRSS). Motor vehicle crashes continue to be a significant public health concern in the United States, resulting in over 38,000 deaths and millions of injuries each year \cite{HighwayTrafficSafetyAdministration2016}. The NHTSA has been collecting crash data since the 1970s using various data collection systems, including the Crash Report Sampling System (CRSS), Fatality Analysis Reporting System (FARS), and Crash Investigation Sampling System (CISS). The Crash Report Sampling System (CRSS) is a compilation of crashes documented by law enforcement officers, encompassing all modes of transportation, including cars, bikes, and pedestrians. The information obtained through the CRSS is analyzed to pinpoint areas of concern regarding road safety, track changes over time, and evaluate the effectiveness of road safety initiatives and regulations \cite{HighwayTrafficSafetyAdministration2016}.
The dataset includes information on various factors related to road crashes, including the location, date and time, type of vehicles involved, driver characteristics, and weather conditions within a period of five years (2016-2020). To be eligible for inclusion in the Comprehensive Roadside Safety Services (CRSS), a traffic accident must meet the following criteria: it must have been reported to the relevant state, involve at least one motor vehicle on a public road, and result in property damage, injury, or death. The CRSS sample is selected from 60 representative regions across the United States, which are chosen based on factors such as geographical location, population density, miles driven, and the frequency of traffic crashes. The sample is drawn randomly from the approximately 6 million traffic accidents reported annually.
\begin{figure}[htbp]
\centerline{\includegraphics[width=0.9\linewidth]{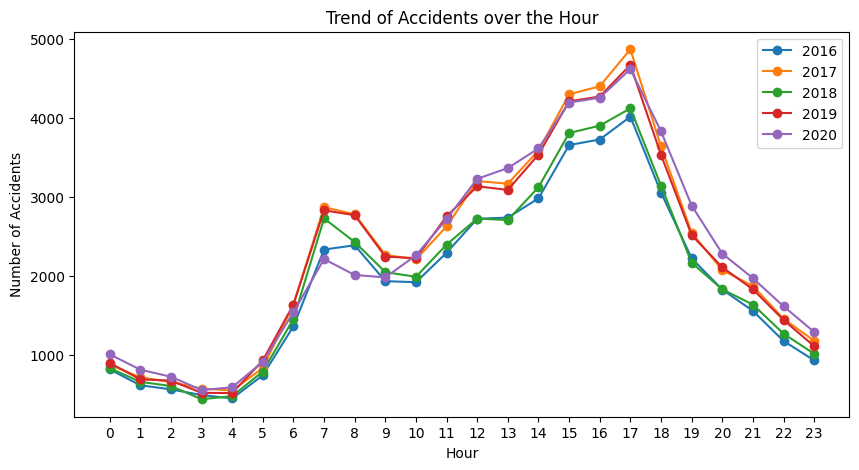}}
\caption{Hourly Trend of Accident (2016-2020).}
\label{Hourtrend}
\end{figure}
Figure \ref{Hourtrend} presents the trend analysis of traffic accidents from 2016 to 2020 by hour. The data shows a consistent increase in accidents every hour, with a noticeable spike between rush hour (7-8 am, 3-6 pm). However, this spike was reduced in 2020. In 2020, traffic accidents were higher than in any other year, particularly from 6 pm to 3 am.
These results suggest that certain times of day may be more susceptible to accidents, which could be due to factors such as increased traffic volume, driver fatigue, or other behavioral issues. 
This highlights the importance of continued attention and efforts toward improving road safety and reducing the number of accidents.
The study by Fors et al. \cite{fors2009night} analyzed night-time traffic in Sweden and found that 25\% of personal injury traffic accidents and 29\% of all fatalities occurred during the late evening (dark hours). To address these issues, physical structures such as street lighting, well-designed roads, traffic signals, and technical solutions such as accident detection systems and weather warning signals can be implemented to reduce the number of traffic accidents during night-time.
Furthermore, Figure \ref{Hourtrend} demonstrates it is essential to consider that human vision is less effective in darkness, highlighting the need for additional measures to ensure road user safety during these times.

\begin{figure}[htbp]
\centerline{\includegraphics[width=0.8\linewidth]{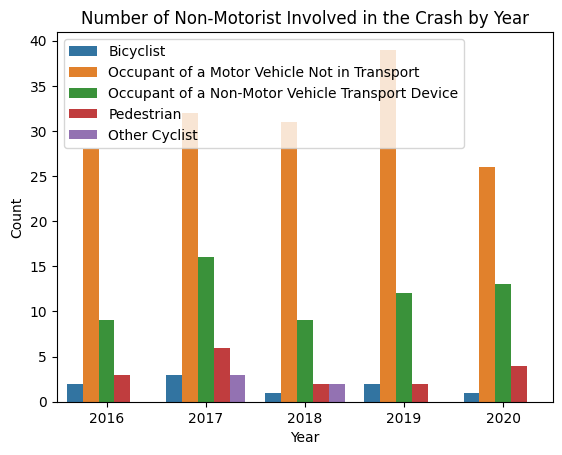}}
\caption{Number of Non-Motorist involved in Car Crash  (2016-2020).}
\label{nonmotorist}
\end{figure}
\begin{figure}[htbp]
\centerline{\includegraphics[width=0.8\linewidth]{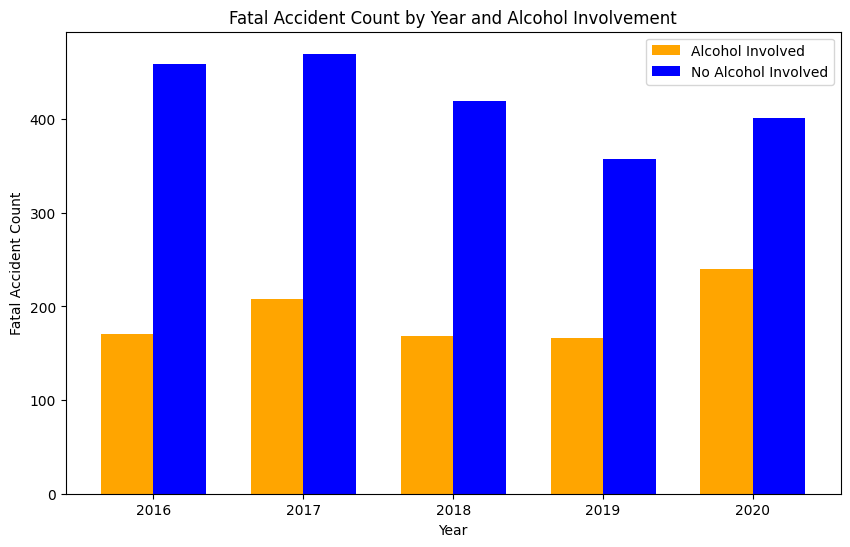}}
\caption{Fatal Accidents with Alcohol Involvement.}
\label{fatalalcohol}
\end{figure}
\begin{figure}[htbp]
\centerline{\includegraphics[width=0.9\linewidth]{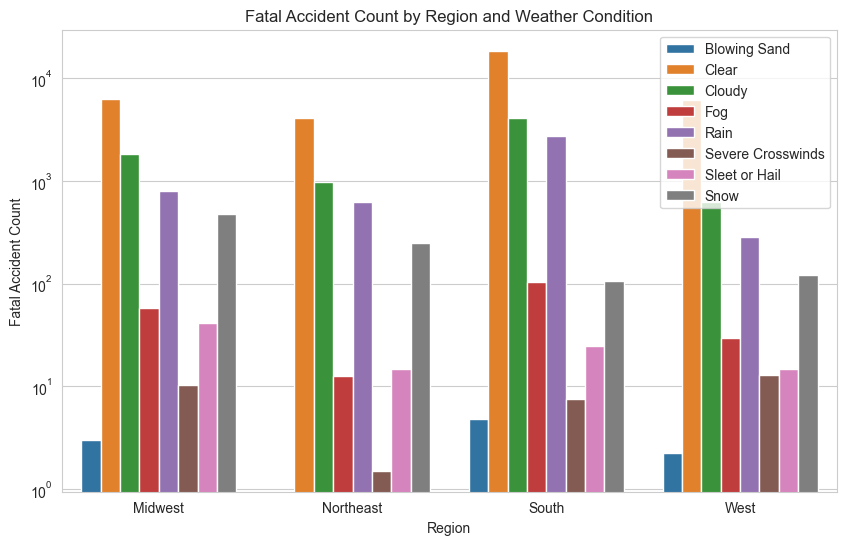}}
\caption{Fatal accident in different regions across the US and weather conditions associated.} 
\label{regions}
\end{figure}

Figure \ref{nonmotorist} shows the number of nonmotorists involved in traffic accidents. The CRSS manual defines nonmotorists as individuals who are not operating a motor vehicle, including pedestrians, cyclists, occupants of motor vehicles that are not in transport, people associated with nonmotorist conveyances such as baby carriages or skateboards, or individuals outside of a traffic way or in a house who are involved in a traffic accident. Most traffic accidents involving nonmotorists involve people in other vehicles not in transport affected by the crash, followed by occupants of non-motor vehicle transport devices. There was a spike in the number of pedestrians involved in traffic accidents in 2017, followed by a significant increase in 2020. On the other hand, the number of cyclists involved in traffic accidents significantly decreased between 2017 and 2019. These findings align with previous research on the vulnerability of nonmotorists in traffic accidents. The study of Justin Tyndall \cite{tyndall2021pedestrian} found that pedestrian fatalities accounted for a disproportionate number of traffic deaths in the united states. One hundred thousand pedestrians death were recorded out of the 741,000  fatalities recorded in the US between 2000 and 2019. Measures such as pedestrian-friendly infrastructure and education campaigns could help to reduce pedestrian accidents \cite{Heinonen2007PedestrianIA}. Bicyclists involved in crashes with motor vehicles are more likely to sustain severe injuries. Another study by National Transportation Safety Board \cite{Bicyclist} found that using bicycle-friendly infrastructure, such as dedicated bike lanes and education campaigns aimed at cyclists and drivers, could help reduce the number of bicycle accidents.
Figure \ref{fatalalcohol} shows traffic fatalities related to alcohol consumption. The NHTSA indicates that the number of traffic accidents involving fatalities related to alcohol consumption was higher in 2020 compared to other years accounting for a 14\% increase from 2019, with 11,654 fatalities recorded in 2020 due to alcohol-impaired driving \cite{nhtsa}. The research of Smailovića et al. \cite{SMAILOVIC2023281} also correlated driving under the influence of alcohol with excessive speed, driving at night-time, and failing to wear seat belts \cite{SMAILOVIC2023281}.
\begin{figure}[htbp]
\centerline{\includegraphics[width=0.8\linewidth]{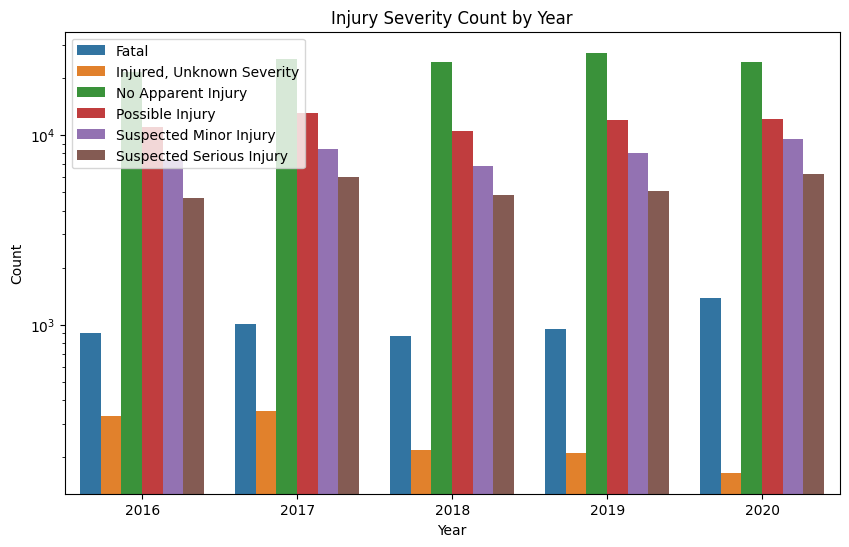}}
\caption{Severity of Injury.}
\label{injury severity}
\end{figure}
There is a noticeable increase in fatal injuries over time, with the highest number recorded in 2020, as shown in Figure \ref{injury severity}. A significant number of suspected injuries and minor injuries were also observed in all five years.
Lin et al.~\cite{lin2022environmental} investigated the environmental factors associated with severe crash injury in Taiwan by linking clinical data with police reports. The study found that drunk driving, early morning driving, flashing signals at intersections, and single-motorcycle crashes were identified as risk factors for severe injury.
Contrastingly, the research indicated that adverse weather conditions do not significantly increase the risk of traffic accidents which is in contrast with the analysis of fatal accidents in different regions in the United States and associated weather conditions shown in Figure \ref{regions}. The authors suggested that a better understanding of traffic patterns during early morning hours and flashing signals at critical intersections could help reduce the number of traffic accidents. Additionally, they found that the time elapsed between the incident and medical care (i.e., the time to the hospital) is a crucial factor in reducing injury severity. This suggests that implementing spontaneous emergency response systems for accident detection could help save more lives and reduce reporting errors \cite{lin2022environmental, Adewopo2022ReviewOA}.
The research of Liu et al. \cite{liu2019exploring} investigated the factors contributing to traffic accidents in low-light conditions in China using a predictive model. Key findings from the research indicate that several factors, such as road conditions, weather conditions, median dividers, and road surface, contribute to fatal traffic accidents. The findings also showed that the probability of fatal accidents is 2.38 times higher at intersections \cite{liu2019exploring}. The analysis in Figure \ref{regions} also illustrates that weather conditions such as cloudy, rain, and snow contribute to fatal accidents in the United States, with the midwest region primarily impacted.

\section{Type of traffic accidents}
\begin{figure}[htbp]
\centerline{\includegraphics[width=0.8\linewidth]{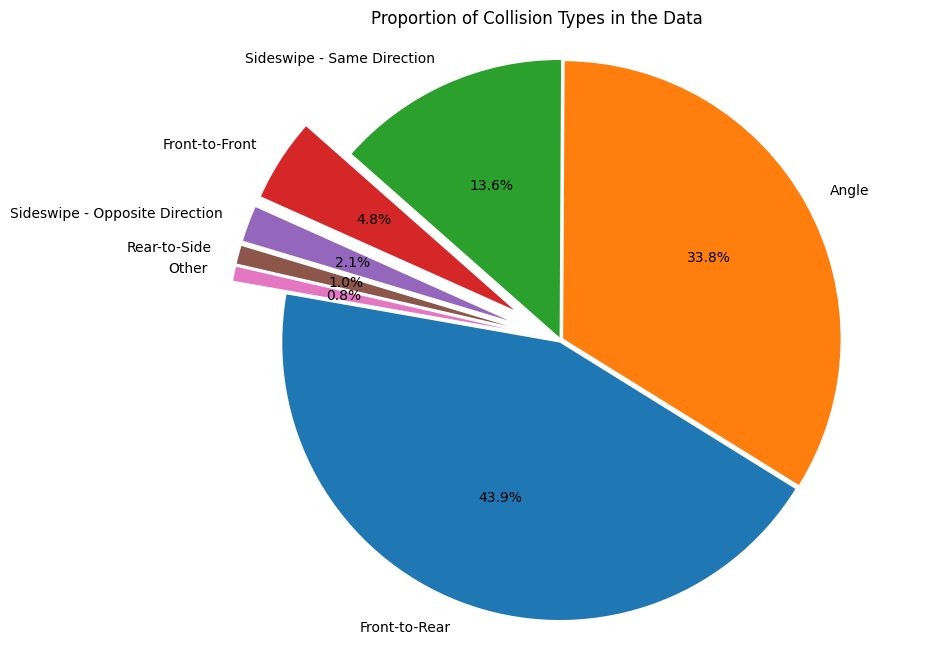}}
\caption{Percentage distribution of different types of collision.} 
\label{collision}
\end{figure}

Traffic accidents is a global challenge causing severe injuries, fatalities, and property damage annually \cite{mohammed2019review}. 
This study addressed only the main types of accidents: rear-end collisions, T-bone or side impact collisions, and front-impact collisions. Figure \ref{collision} showcase the percentage distribution of collision types with Fron-to-Rear collision accounting for 43.9\% of traffic accidents, 33.8\% accounts for angle, and 13.6\% accounting for same-direction traffic accidents. For effective interventions to reduce their frequency and severity and to develop effective strategies that can reduce their occurrence and severity and improve road safety, it is essential to understand the factors contributing to these accidents.

\subsection{Rear-end collision} 
According to the National Highway Traffic Safety Administrator, Rear-end collisions are the most common type of traffic accident, accounting for 29\% of all crashes and causing significant injuries and fatalities annually \cite{Hs2007AnalysesOR}. Rear-end collisions can result in injuries, from minor whiplash to severe head and spinal injuries and even fatalities. Several factors contribute to rear-end collisions, including distracted driving, following too closely, sudden stops, and poor weather conditions. One of the leading causes of rear-end collisions is driver distraction, such as texting or talking on the phone while driving. Additionally, distracted drivers cannot react quickly enough to avoid a collision in the event of sudden stops or slow down of vehicles ahead \cite{eboli2020factors}. Efforts to address this issue had limited success, with the center high-mounted stop lamp (CHMSL) being one of the most notable initiatives launched in 1986. Although the CHMSL has reduced rear-end collisions by 4\%, further improvement is still needed. 
To prevent rear-end collisions, drivers can manually employ safety measures such as increasing the distance between vehicles, reducing speed in poor weather conditions, and avoiding distractions while driving. Advanced driver assistance systems (ADAS) can be installed in vehicles to help drivers avoid collisions. These systems use sensors and cameras to detect potential collisions and warn the driver \cite{kukkala2018advanced}. Developing new intelligent transportation systems (ITS) can help prevent rear-end collisions. These systems use connected vehicle technology to allow vehicles to communicate with each other (V2X communication) and with infrastructure (Collision avoidance, lane detection), providing drivers with real-time information about road conditions and potential hazards (indoor monitoring) 
\cite{kukkala2018advanced}.

\subsection{T-Bone Collision}
T-bone accidents are also often referred to as side-impact collisions. T-bone accidents are likely to be severe injuries based on the lack of structural barrier between driver and passengers engaged in such events depending on the advanced safety features available in the car. 
T-bone collisions can result in serious injuries or fatalities for occupants of the struck vehicle due to less protection on vehicle sides \cite{mohammed2019review}. T-bone collisions account for approximately 13\% of all passenger vehicle occupant fatalities in the United States. In addition, side impact collisions are the second most common type of fatal collision for passenger vehicle occupants, after frontal crashes \cite{nhtsa}. 
Technological advancements have also been developed to mitigate the severity of side-impact collisions. For example, side airbags are now standard in many new vehicles and can provide additional protection to occupants in a side-impact collision. Moreover, some vehicles are equipped with advanced safety features such as automatic emergency braking and lane departure warning systems, which can help prevent collisions from occurring. 
Eboli et al. \cite{eboli2020factors}, found some correlation between road structures, driver factors, and types of accident collision through the analysis of road accident type. Surface conditions, such as dry road surfaces, reduce the probability of a front/side collision in serious accidents. Driver-related factors, such as having a car license, increase the probability of a front/side collision in serious accidents. Furthermore, environmental factors, such as sunny weather, increase the probability of a front/side collision. The authors also pointed out that driver-related factors play a more important role in the probability of a front/side collision \cite{eboli2020factors}.

\subsection{Frontal impact accident}
Frontal impact accident accounts for the most inter-vehicular accident with severe injuries and deaths; it is the frontal impact with 35\% severity impact \cite{jirovsky2015classification}. 
Figure \ref{collision} illustrates that only 4.8\% of traffic accidents are due to frontal collisions. These collisions often result in serious injuries or fatalities due to the force of impact. Front-end collisions can occur for a variety of reasons, including driver error, speeding, distracted driving, and poor weather or road conditions. In some cases, faulty vehicle equipment or manufacturing defects may also contribute to front-end collisions. 
Efforts made to reduce the incidence and severity of front-end collision includes using safety features such as airbags, crumple zones, and seat belts, as well as developing advanced driver assistance systems (ADAS) that can alert drivers to potential collisions and even take action to avoid them. The existing forward collision warning (FCW) systems based on kinematic or perceptual parameters have some drawbacks in warning performance due to poor adaptability and ineffectiveness \cite{sheoreysensor}. To address these problems, machine learning and deep learning algorithms have been proposed. However, these models lack consideration for multi-staged warnings, which could distract or startle the driver. A light gradient boosting machine (LGBM) learning algorithm was used to develop a multi-staged FCW model, which was evaluated using a driving simulator by twenty drivers \cite{ma2022adaptive}. The study found that the front vehicle acceleration, time-to-collision (TTC), and relative speed strongly affected the warning stages from the proposed model. The authors aim to develop LGBM for developing FCW models that could improve warning performance while considering multi-staged warnings~\cite{ma2022adaptive}.
The study of Jirovsky et al.~\cite{jirovsky2015classification} employed a new approach for determining the probability of collision between two vehicles by defining a 2-D reaction space, which describes all possible positions of the vehicles in the future. This approach enables mitigation of collision by exploring alternative causes of action such as changing direction in addition to braking \cite{jirovsky2015classification}.
\begin{figure}[htbp]
\centerline{\includegraphics[width=0.8\linewidth]{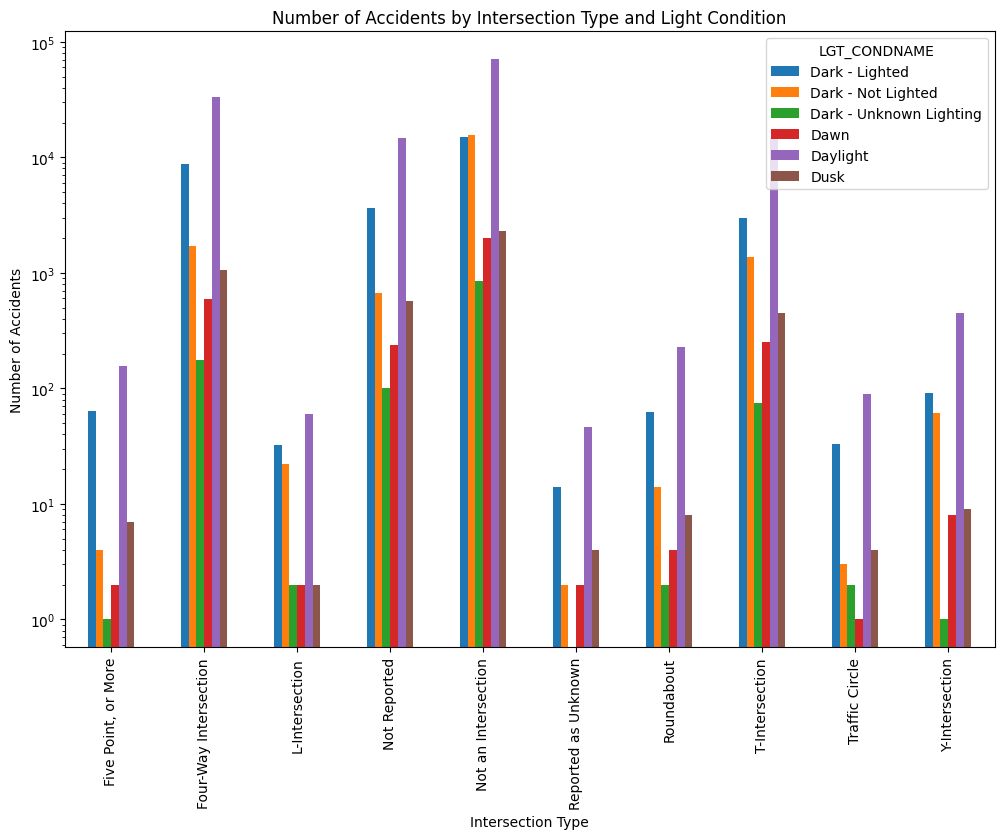}}
\caption{Accidents by Intersection Type and Light Condition Associated.} 
\label{intersection}
\end{figure}
Figure \ref{intersection} shows the analysis of traffic accidents at different intersections and the light conditions. Light condition plays a crucial role in traffic safety, and the data indicates that accidents are more common at intersections, with four-way intersections being hazardous. Whether lighted or not, the number of accidents in dark conditions is high across all intersections. The data also shows that accidents in dark-light conditions are more frequent at T-intersections and four-way intersections.
Adequate road lighting is crucial for ensuring the safety of drivers during nighttime driving. The study by Alharbi et al. \cite{alharbi2023performance} on the performance appraisal of urban lighting systems aligns with the findings in Figure \ref{intersection}. The researchers found that electronic billboard positioning, oncoming vehicle lights, and poor lighting conditions during inclement weather, particularly dust, are significant factors affecting the performance of urban street lighting systems (USLSs) as perceived by road users. The research by Bridger et al. \cite{bridger2012lighting} reported that modern LED lighting is a disruptive technology and that decreasing nighttime fatalities and injuries due to modern road lighting has significant cost benefits. On the other hand, the study by Marchant et al.~\cite{marchant2022determine} found limited evidence to support road safety improvement through relighting traffic lights with white lamps in the UK. 

\section{Methodology}
The transportation industry has experienced significant growth in implementing computer vision systems. These systems have proved to be instrumental in enhancing traffic safety, mitigating traffic congestion, and improving transportation efficiency \cite{yu2021deep}. Among the various computer vision systems, action recognition has been identified as a crucial area of focus in the transportation industry, specifically in autonomous transportation and road traffic safety management. AR systems have the potential to significantly improve the safety and reliability of transportation systems in smart cities by providing real-time traffic management and accident detection \cite{You2020}. Adopting autonomous transportation systems in smart cities will immensely benefit from cutting-edge systems such as radar detection, wrong-way detection, license plate recognition, and road construction navigation, all of which can be enhanced by developing advanced algorithms and technologies. The continued advancement of action recognition technologies in the transportation industry is essential to improve efficiency and safety.
Huang et al. \cite{huang2020intelligent} propose an intelligent intersection system that utilizes two-stream convolutional networks to detect near-accidents in real-time traffic video. The system combines spatial and temporal information to improve the accuracy of detecting near accidents. The experimental result of the proposed system reported high accuracy in the real-time detection of near-accidents with the potential to enhance road safety by alerting drivers and reducing the occurrence of accidents at intersections. Zhao et al. \cite{zhao2017temporal} proposed a novel approach in temporal action detection by predicting actions' start and end times in untrimmed videos. Structured Segment Networks (SSN) use a two-stage process in generating segment proposals, including the start time, event course, and end time of action with appropriate action tags. 
Similarly, Lee et al. \cite{Lee2020} proposed weakly-supervised temporal action localization to detect intervals of action instances with only video-level action labels and separate background frames that do not contain an action class. The proposed model was tested on the Thumos dataset using the I3D architecture, the feature extraction process involved splitting videos into multi-frame non-overlapping segments to mitigate videos with varied lengths. 
The model was optimized using three losses: action classification losses, uncertainty modeling loss, and background entropy loss. This method can be used to improve action recognition models in real-world scenarios and improve such models' accuracy.
\section{Proposed Framework}
\begin{figure}[htbp]
\centerline{\includegraphics[width=0.8\linewidth]{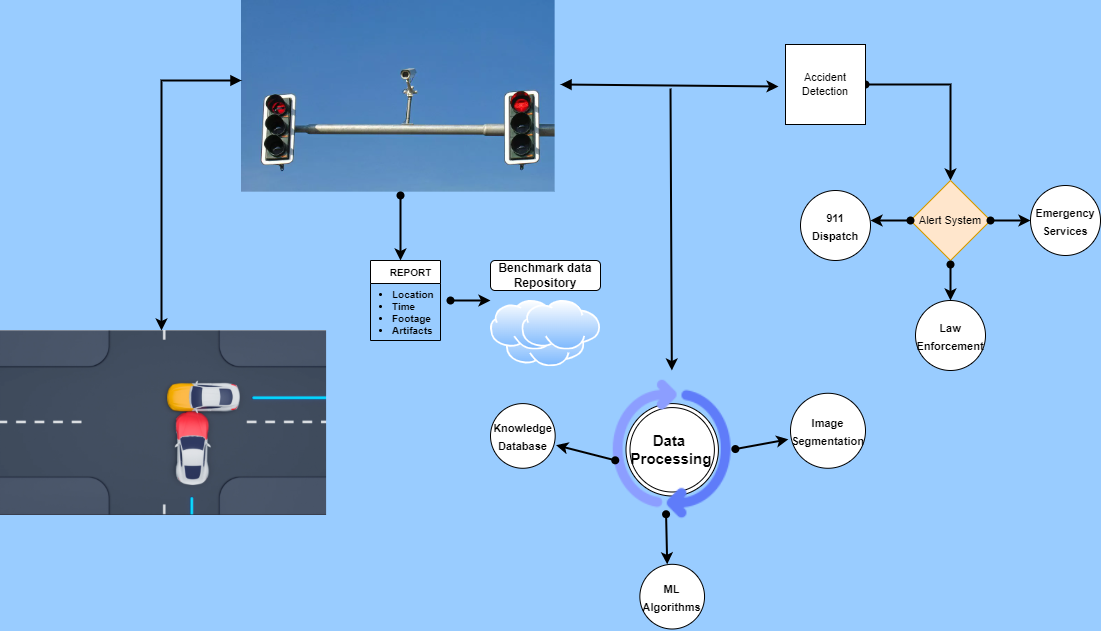}}
\caption{Proposed Framework for Autonomous Accident Detection System.} 
\label{framework}
\end{figure}
The proposed accident detection framework for autonomous transportation focuses on three key elements: Traffic Camera Utilization, Data Processing, and Accident Detection. These components work synergistically to create a comprehensive and efficient system for identifying and responding to traffic accidents.
\begin{itemize}
    \item \textbf{Traffic Camera Utilization}: Leveraging existing traffic surveillance cameras, we repurpose them to perform more sophisticated tasks such as accident detection and monitoring traffic violations like red light running. These cameras continuously capture valuable information such as location, time of the accident, accident footage, and other artifacts. The collected data is then transferred to a central repository for further analysis and processing.

    \item \textbf{Data Processing}:
The data extracted from the traffic cameras undergo a series of processing steps, including image segmentation and application of machine learning algorithms. These algorithms use the knowledge database to analyze the data and identify accident-related patterns. The output generated from data processing is then sent to the Accident Detection System for further evaluation and decision-making.
    
    \item \textbf{Accident Detection and Alert System}:
 The Accident Detection System comprises two integral components that revolve around a decision alert system. These components work together to respond to detected accidents promptly. Upon identifying an accident, the alert system initiates the following actions:
    \begin{enumerate}
        \item \textit{The 911 Dispatch Call:} The system automatically places a call to the 911 emergency dispatch center, providing essential information such as location, time, and nature of the accident. This ensures that first responders are notified promptly and can reach the scene as quickly as possible.
        \item \textit{Law Enforcement Communication:} Lastly, the system alerts law enforcement agencies about the accident, enabling them to manage traffic flow, investigate the incident, and take appropriate legal action as needed.
    \end{enumerate} 
\end{itemize}
This paper presents a novel framework for accident detection in autonomous transportation systems that harnesses the power of traffic cameras, data processing, and machine learning algorithms. By automating the alert system and coordinating necessary emergency services, this framework can improve road traffic safety and reduce the severity of traffic accidents. Previous studies have approached traffic accident detection from different angles, including clustering, fuzzy logic, K-Means, and linear regression. However, these methods often create computational and cost constraints because the developed model may need to be more generalizable to other areas or traffic routes not included in the data collection. 
The proposed framework builds upon the DSTGCN framework proposed by Le et al.~\cite{Yu}, which is efficient in predicting accidents at the road level by modeling the spatial correlations and temporal dependencies of the road structure. By leveraging the power of machine learning algorithms and incorporating data from traffic cameras, this framework can detect accidents in real time and respond quickly and efficiently to minimize the severity of accidents. Our research presents a promising solution for accident detection in autonomous transportation systems and contributes to developing a safer and more efficient transportation system.

\section{Conclusion and discussion}
Our analysis of traffic accidents from 2016 to 2020 highlights the continued proliferation of traffic accidents and the urgent need for practical solutions to reduce the number of accidents and improve road safety. The findings of this study indicate that the number of traffic accidents has been increasing over the years, with a noticeable spike in the number of accidents in the early morning and late evening. We also observed high spikes at T-intersections and four-way intersections with different light and weather conditions. Additionally, the number of accidents involving non-motorists, such as pedestrians and cyclists, has increased, highlighting the need for further research and quick intervention. 
This also underpins the importance of developing effective strategies for reducing the number of fatal accidents and improving road safety. Our analysis of traffic accidents is consistent with the research findings of other studies on the impact of various factors on traffic safety, including light conditions, weather conditions, road design, and driver behavior. Advanced technologies, such as autonomous accident detection systems, can play a crucial role in reducing the number of accidents and improving road safety by providing real-time data on the severity of accidents and improving the speed and efficiency of emergency response, which has proven to reduce accident severity.

Furthermore, the analysis of fatalities related to alcohol consumption and the severity of injuries highlights the need for continued attention to reducing driving under the influence of alcohol and improving emergency response. 
Our proposed framework utilizes Action Recognition (AR) for accident detection, which can provide a valuable tool for detecting accidents when they occur and providing real-time updates to other motorists. In light of the growing reliance on efficient systems, automated tools, and smart city technologies, the need for reliable and accurate accident detection systems will likely have an upward trend. Technological progress and logical reasoning in developing innovative solutions utilizing existing technologies are imperative to enhance public safety and minimize traffic accidents. With the aid of these technologies, law enforcement officials will be able to investigate accident scenes and conduct thorough investigations. 
In conclusion, this study highlights the importance of continued research and attention to improving road safety and reducing the number of traffic accidents. Advanced technologies, such as autonomous accident detection systems, can be crucial in this effort. However, further research is needed to fully understand the reasons behind the trends in traffic accidents and determine the most effective strategies for reducing the number of accidents. The elapsed time between an accident and medical emergency attendants is also a crucial factor in reducing the severity of accidents, and the proposed framework provides an opportunity to understand the best strategies for improving emergency response times.



\bibliography{main}\bibliographystyle{ieeetr}

\end{document}